\documentclass{article}
\usepackage{spconf,amsmath,epsfig,hyperref}

\usepackage{subcaption}
\usepackage{hyperref}

\def\cpar{\hss\egroup\line\bgroup\hss}

\title{Fast model inference and training on-board of Satellites}
\name{Vít Růžička\textsuperscript{1,2},
    Gonzalo Mateo-García\textsuperscript{2,3}, 
    Chris Bridges\textsuperscript{4},\vspace{-2.2ex}
}
\address{\textit{Chris Brunskill}\textsuperscript{6}, 
    \textit{Cormac Purcell}\textsuperscript{2,5}, 
    \textit{Nicolas Longépé}\textsuperscript{7}, 
    \textit{Andrew Markham}\textsuperscript{1} 
    \thanks{Correspondence to: \texttt{vit.ruzicka@cs.ox.ac.uk}}
    \thanks{This work has been funded by ESA Cognitive Cloud Computing in Space initiative. G.M.-G. has been partially supported by the Spanish Ministry of Science and Innovation (project PID2019-109026RB-I00 funded by MCIN/AEI/10.13039/501100011033) and the European Social Fund.}\\ \\
    \textsuperscript{1} University of Oxford, 
    \textsuperscript{2} Trillium Technologies, 
    \textsuperscript{3} University of Valencia, \\
    \textsuperscript{4} University of Surrey, 
    \textsuperscript{5} University of New South Wales, 
    \textsuperscript{6} D-Orbit
    \textsuperscript{7} European Space Agency
}

\begin{document}
%
\maketitle
\begin{abstract}
Artificial intelligence onboard satellites has the potential to reduce data transmission requirements, enable real-time decision-making and collaboration within constellations. This study deploys a lightweight foundational model called RaVAEn on D-Orbit's ION SCV004 satellite. RaVAEn is a variational auto-encoder (VAE) that generates compressed latent vectors from small image tiles, enabling several downstream tasks. In this work we demonstrate the reliable use of RaVAEn onboard a satellite, achieving an encoding time of 0.110s for tiles of a 4.8x4.8 km² area. In addition, we showcase fast few-shot training onboard a satellite using the latent representation of data. We compare the deployment of the model on the on-board CPU and on the available Myriad vision processing unit (VPU) accelerator. To our knowledge, this work shows for the first time the deployment of a multi-task model on-board a CubeSat and the on-board training of a machine learning model.
\end{abstract}
\begin{keywords}
Training on-board, AI on satellites, efficient neural network models
\end{keywords}
\section{Introduction}
\label{sec:intro}

Onboard data processing plays a crucial role in maximizing the potential of Earth-observation (EO) satellites. With the significant increase in EO data volume, it is essential to have efficient and intelligent processing capabilities directly onboard the satellites~\cite{Furano_towards_AI_2020}. By leveraging onboard data processing, satellites can perform advanced analysis and make critical decisions on the acquired data in real-time. Several applications have already been tested in demonstration missions, such as prioritizing imaging targets~\cite{ruzicka2022unsupervised}, discarding non-usable images~\cite{giuffrida_phisat-1_2021}, identifying events of interest~\cite{fanizza_ship_2022,kplabs_onboard_2021,ruzicka_starcop_preprint} or compressing the output to rapidly transmit relevant information to the ground~\cite{mateo2022orbit,hs_inf}.

\begin{figure}[h]
    \centering
    \includegraphics[width=.92\linewidth]{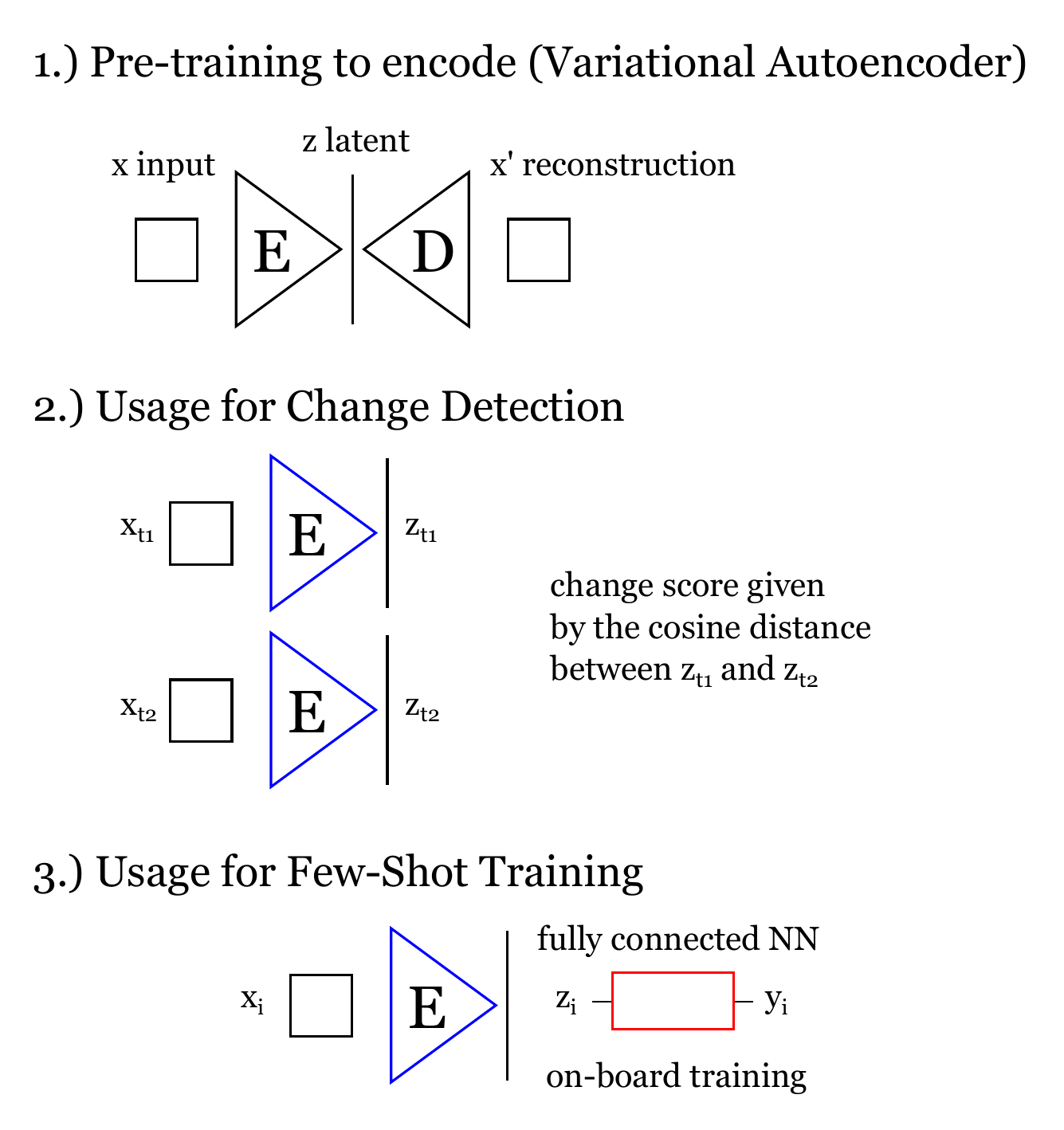}
    \caption{Illustration of the different uses of the variational autoencoder model RaVAEn. Pre-training (1.) conducted in prior work \cite{ruzicka2022unsupervised}. In this paper, we measure the inference time of the encoder network (denoted as $E$ and marked with blue colour as with frozen weights) for the task of unsupervised change detection (2.), and time needed for on-board training (3.) of a small fully-connected neural network classification model.}
    \label{fig:illustration}
\end{figure}

In this work we go one step further and deploy a lightweight foundational model called RaVAEn on-board of D-Orbit’s ION SCV004 satellite demonstration mission. RaVAEn is a variational auto-encoder (VAE) model that generates latent vectors from small tiles of the original image. These latents can be used for several other tasks, such as change detection, as shown in our previous publication in the context of disaster response~\cite{ruzicka2022unsupervised}, or as features extracted to train other downstream models. 

We show that this model can be reliably used with the compute available directly on-board of the D-Orbit’s ION-SCV 004 satellite. 
In addition, we also demonstrate to the best of our knowledge the world’s first fast and efficient few-shot training on-board of a satellite using the latent representation of the data. To this end, we use the learned encoder of the VAE model to represent tiles of $32\times32$ pixels with $4$ bands as $128$-dimensional latent vectors. We then train a lightweight classification model using these latent vectors as inputs in a few-shot learning manner. Good representation of the Sentinel-2 data is required for training with only limited number of samples. As a demonstration task, we select cloud detection: in this context the decision if a tile contains clouds or not. This task is relevant for on-board data processing as it has been previously used to select which image acquisitions are downlinked to the ground station and which are to be ignored \cite{wagstaff2017cloud, giuffrida_phisat-1_2021}. 




To summarise, our contributions are:
\begin{itemize}
    \item Measuring the inference time of the RaVAEn model encoder on three different compute regimes available: Myriad VPU, or CPU with Pytorch, or OpenVino libraries.
    \item Demonstrating few-shot training directly on-board of a satellite for a task of cloud detection, as a motivation for future on-board auto-calibration tasks. To the best of our knowledge, this is the world’s first case of on-board few-shot training on-board of a satellite.
\end{itemize}

We release the used code in a GitHub repository at \href{https://github.com/previtus/RaVAEn-unibap-dorbit}{{previtus/RaVAEn-unibap-dorbit}}

\section{Methodology}

The RaVAEn uses a VAE model \cite{kingma2013auto} trained on Sentinel-2 L1C data from the WorldFloods dataset \cite{mateo2021towards}. The VAE model consists of an encoder network that reduces the dimensionality of the input data into a latent vector, and of a decoder network that has to reconstruct the original data from this compressed representation. The learned encoding space has to learn an informative representation of the data. This can be leveraged for unsupervised change detection, where, instead comparing changes in the pixel space (which can be noisy due to a wide array of effects), we compare the data representations in the latent space. This approach was evaluated on an annotated dataset of disaster events containing samples with landslides, floods, hurricanes and fires in \cite{ruzicka2022unsupervised}. In this follow-up paper, we are interested in the inference time on real satellites and in exploring new tasks that this architecture allows us to do directly on-board. Namely, we are interested in the inference time required to process Sentinel-2 data by the encoder network of the RaVAEn model. We note, that the original architecture was designed with a requirements for fast inference in mind, and the fastest model was chosen from several variants.

Additionally, we explore training the mdoel on-board satellites, but instead of using the full dimensionality of the input data, we leverage the general pre-trained encoder of the RaVAEn model. We train a tiny classification model on the encoded latent vectors directly on-board of the satellite. The general VAE encoder is capable of efficient data representation, which we can use for few-shot learning. The suitability of few-shot learning has been highlighted by \cite{derksen2021few}. Importantly, the resulting training process is faster and requires fewer labels than would be required when training the VAE model from scratch. From the dataset of Sentinel-2 L1C images, we select $1305$ tiles (cloudy and not-cloudy) for the training dataset and use the non-overlapping remainder of the data for evaluation. We note that other approaches frame cloud detection as semantic segmentation task, however as a demo task, per tile classification is sufficient and can still provide us with a rough estimation of the percentage of clouds in a scene. 

\textbf{Hardware} The ION-SCV 004 satellite has the following relevant specifications: a quad-core x86 64-bit CPU processor, Intel Movidius Myriad X VPU and 2GB RAM. Similar configuration was used in the work of \cite{mateo2022orbit}. We note, that for smaller CubeSats, model training can be offloaded to these satellites. 

\section{Results}

\begin{figure}
    \centering
    \includegraphics[width=8.5cm]{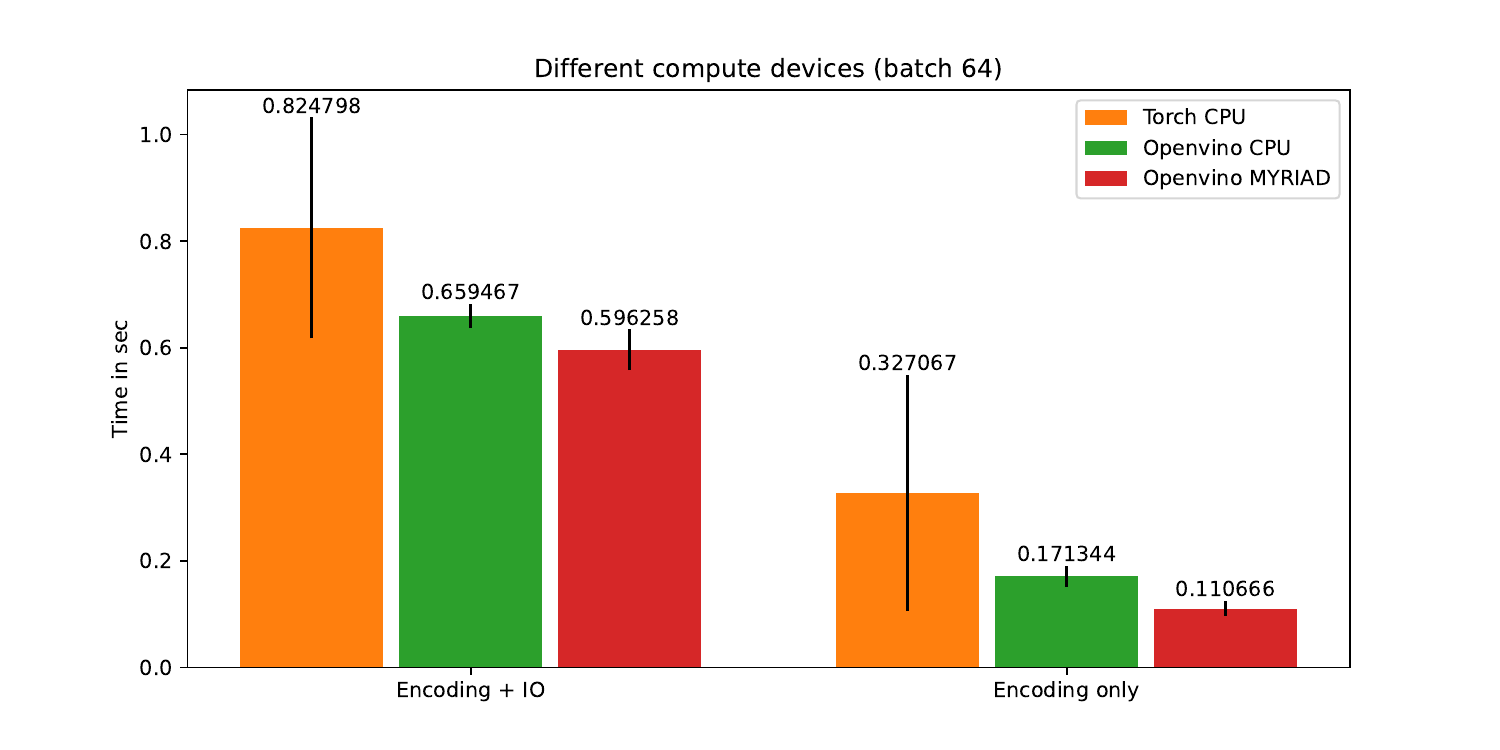}
    \caption{Timed measuments model inference, using the RaVAEn encoder network with different devices available available on the satellite.}
    \label{fig:timings_inference}
\end{figure}

\begin{figure}[ht]%
    \centering
    \subfloat[\centering CPU with Pytorch]{{\includegraphics[width=8.5cm]{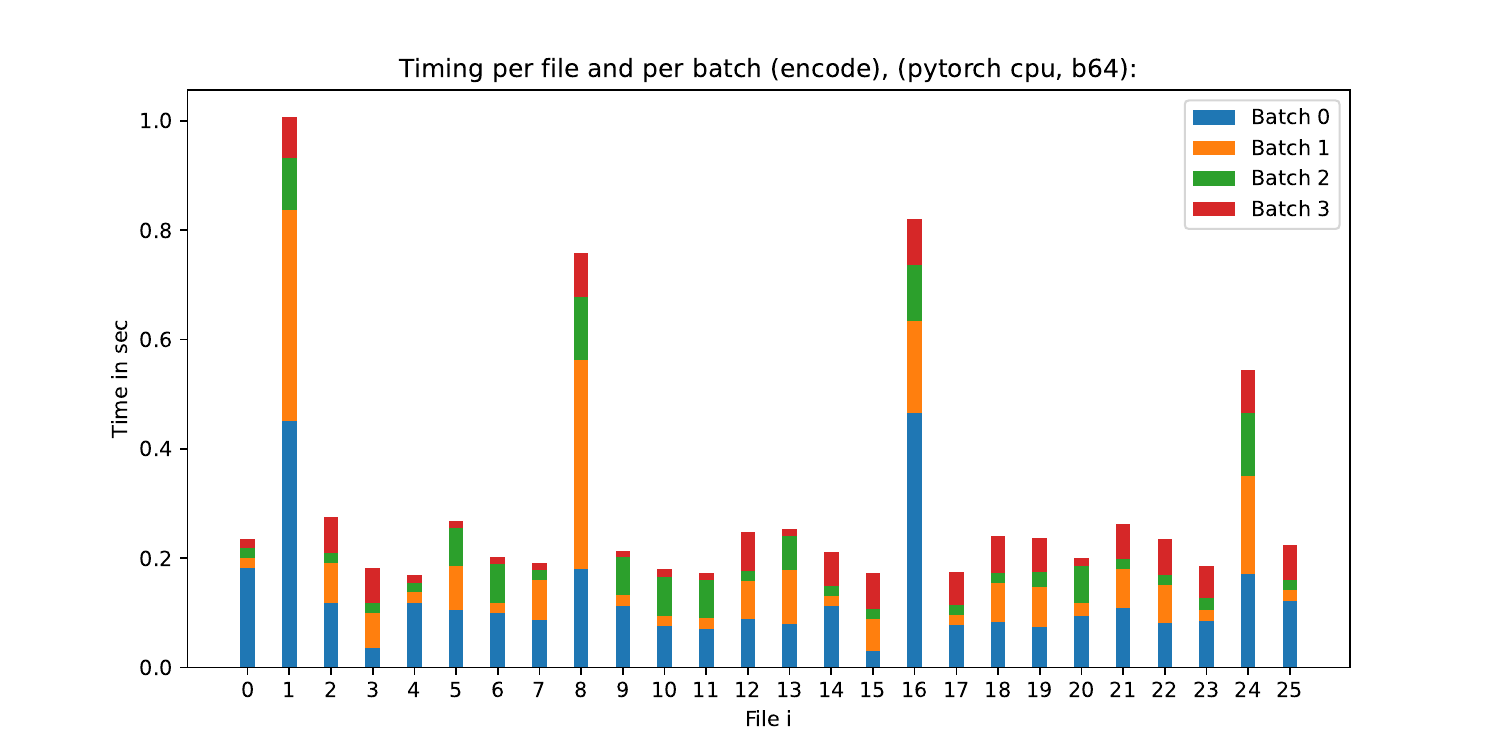} }}%
    \qquad
    \subfloat[\centering Myriad VPU with OpenVino]{{\includegraphics[width=8.5cm]{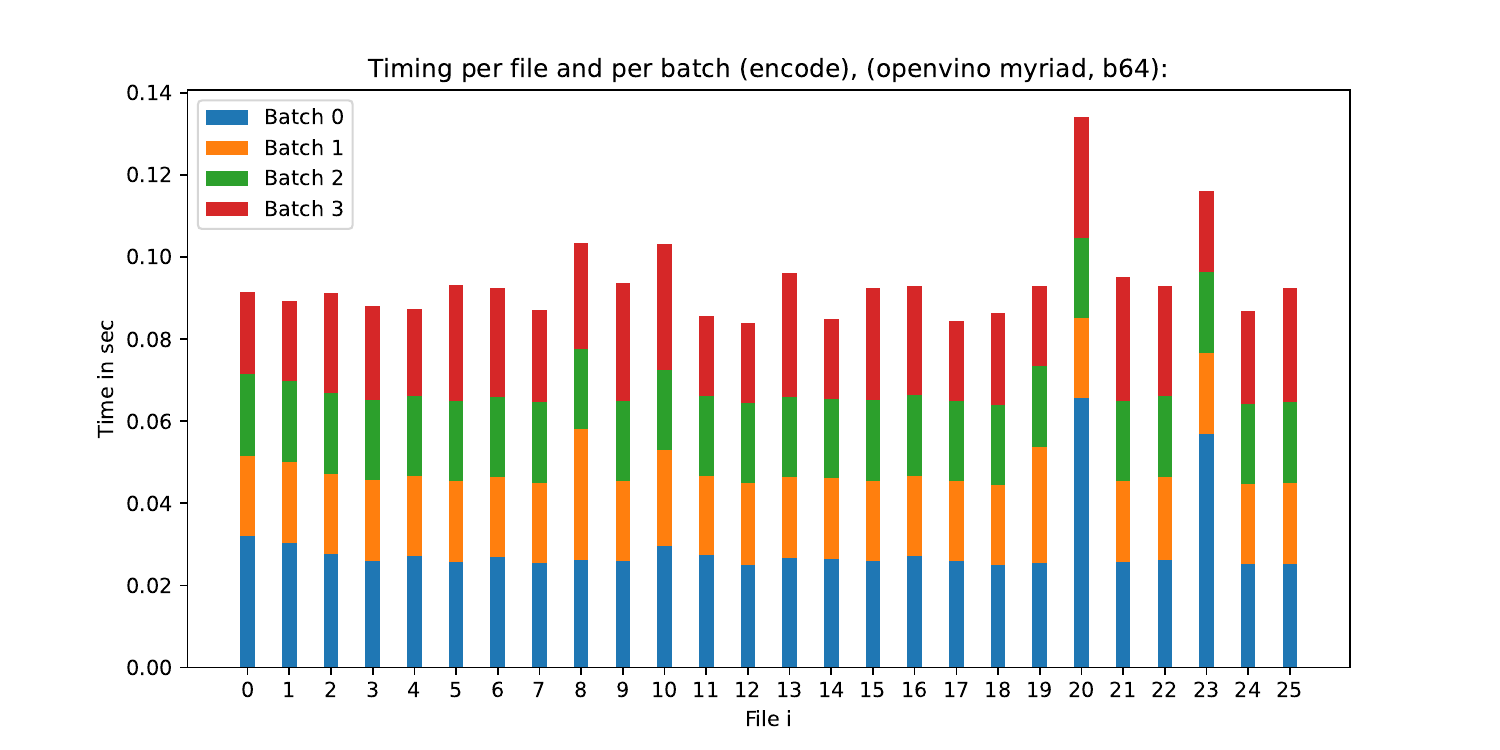} }}%
    \caption{Detailed view into the individual batches used in the RaVAEn encoder, showing delays in the CPU regime.}%
    \label{fig:detailed}%
\end{figure}

\textbf{Model inference with RaVAEn} We measure the time required to load, encode and compare all the tiles from a sequential dataset of Sentinel-2 images. In Figure~\ref{fig:timings_inference} we show the average inference time of encoding one file (consisting of $225$ tiles) with a batch size of $64$ tiles, using the RGB+NIR bands. We observe that the deployment of the model on the Myriad VPU with OpenVino offers the fastest encoding time. Furthermore, when inspecting the individual encoding times per batch of each file in Figure~\ref{fig:detailed}b, we see that the Myriad VPU is also more robust to slowdowns, which occur when using the CPU with PyTorch (Fig.~\ref{fig:detailed}a). The relatively slow loading and tiling of the images can be speed up if we process data with delay and parallelisation (as shown in \cite{ruuvzivcka2018fast}).

\textbf{Training models on-board of satellites} We measure training times and also the performance on the downstream task. On the demonstration task of cloud detection, we get an AUPRC score of $0.979$. With a confidence threshold of $0.5$, we get recall of $0.946$, precision of $0.967$ and a F1 score of $0.956$. We note that this task serves only as a demo, as we are mostly interested in the timing of the entire training process.

\begin{figure}
    \centering
    \includegraphics[width=5.5cm]{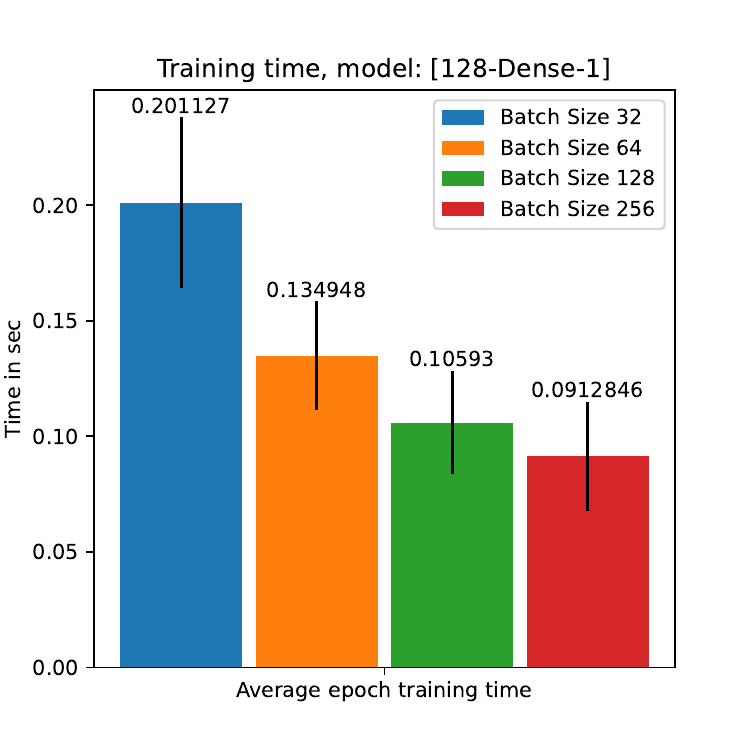}
    \caption{Training a classification model on top of the encoder network, timing while changing the used batch size.}
    \label{fig:timings_training}
\end{figure}

On Figure~\ref{fig:timings_training}, we see the average time measurements for each epoch when using different batch sizes, and when training a tiny one-layer binary classification model with $129$ trainable parameters. With the batch size of $256$, one epoch takes on average only $0.091s$ to train.


\section{Discussion and conclusion}

The domain of AI on-board of satellites is unique in comparison with the rest of computer vision research, as the remote sensing data usually undergoes heavy post-processing steps after it has been down streamed to the ground station. On-board processing and training has to, on the other hand, deal with the near-raw data capture, which poses unique opportunities such as re-training existing models with newly observed data - a challenge identified by \cite{furano2020ai}. 

In this work, we demonstrate the possibilities of training directly on-board of the satellite, which is of interest for future self-calibration tasks. Training on-board is feasible in scenarios where we obtain both raw measurements of the scene and a reliable annotation. This can occur in cases, where the instrument carries samples with known ground truth labels \cite{bell2017mars}, in cases when we are orbiting around a known location with known state of the observed data, or in cases where another satellite provides us with labels due to having access to more powerful or more reliable instruments (as can be the case when working with a mixture of multispectral and hyperspectral as hypothesized in \cite{ruzicka_starcop_preprint}).

In comparison with \cite{furano2020ai}, which suggests uplinking updated versions of model weights from the ground stations, we propose training on-board as a new approach for adapting AI models in the space. This may be more beneficial for security reasons, and in communication constrained environments, where collection possibilities of new data outweights the transmission limitations. We consider these scenarios as exciting new opportunities to explore for increasing autonomy of satellites deployed around the Earth and in deep space.



\bibliographystyle{IEEEbib}
\bibliography{refs}

\end{document}